\pdfoutput=1
\documentclass[10pt,twocolumn,letterpaper]{article}

\usepackage{cvpr}
\usepackage{times}
\usepackage{epsfig}
\usepackage{epstopdf}
\usepackage{graphicx}
\usepackage{amsmath}
\usepackage{amssymb}
\usepackage{dsfont}
\usepackage{color}
\usepackage{wrapfig}

\usepackage[pagebackref=true,letterpaper=true,colorlinks,bookmarks=false]{hyperref}

\cvprfinalcopy 

\ifcvprfinal\fi
\begin{document}

\title{Realtime Multi-Person 2D Pose Estimation using Part Affinity Fields \thanks{Video result: \url{https://youtu.be/pW6nZXeWlGM}}}


\author{Zhe Cao\quad\quad Tomas Simon \quad\quad Shih-En Wei \quad\quad Yaser Sheikh\\
The Robotics Institute, Carnegie Mellon University\\
{\tt\small \{zhecao,shihenw\}@cmu.edu \quad\quad \{tsimon,yaser\}@cs.cmu.edu}
}

\maketitle

\begin{abstract}
We present an approach to efficiently detect the 2D pose of multiple people in an image. The approach uses a nonparametric representation, which we refer to as Part Affinity Fields (PAFs), to learn to associate body parts with individuals in the image. The architecture encodes global context, allowing a greedy bottom-up parsing step that maintains high accuracy while achieving realtime performance, irrespective of the number of people in the image. The architecture is designed to jointly learn part locations and their association via two branches of the same sequential prediction process. Our method placed first in the inaugural COCO 2016 keypoints challenge, and significantly exceeds the previous state-of-the-art result on the MPII Multi-Person benchmark, both in performance and efficiency.

\end{abstract}

\section{Introduction}

Human 2D pose estimation---the problem of localizing anatomical keypoints or ``parts"---has largely focused on finding body parts of \emph{individuals}~\cite{fh2005pictorial,Andriluka2010,Andriluka2009,pishchulin2013poselet,yang2013articulated,johnson2010clustered,ramanan2005strike,wei2016convolutional,bulat2016human,Ramakrishna2014posemachines}. Inferring the pose of multiple people in images, especially socially engaged individuals, presents a unique set of challenges. First, each image may contain an unknown number of people that can occur at any position or scale. Second, interactions between people induce complex spatial interference, due to contact, occlusion, and limb articulations, making association of parts difficult. Third, runtime complexity tends to grow with the number of people in the image, making realtime performance a challenge.

\begin{figure}[t]
    \centering
    \includegraphics[width=1\linewidth]{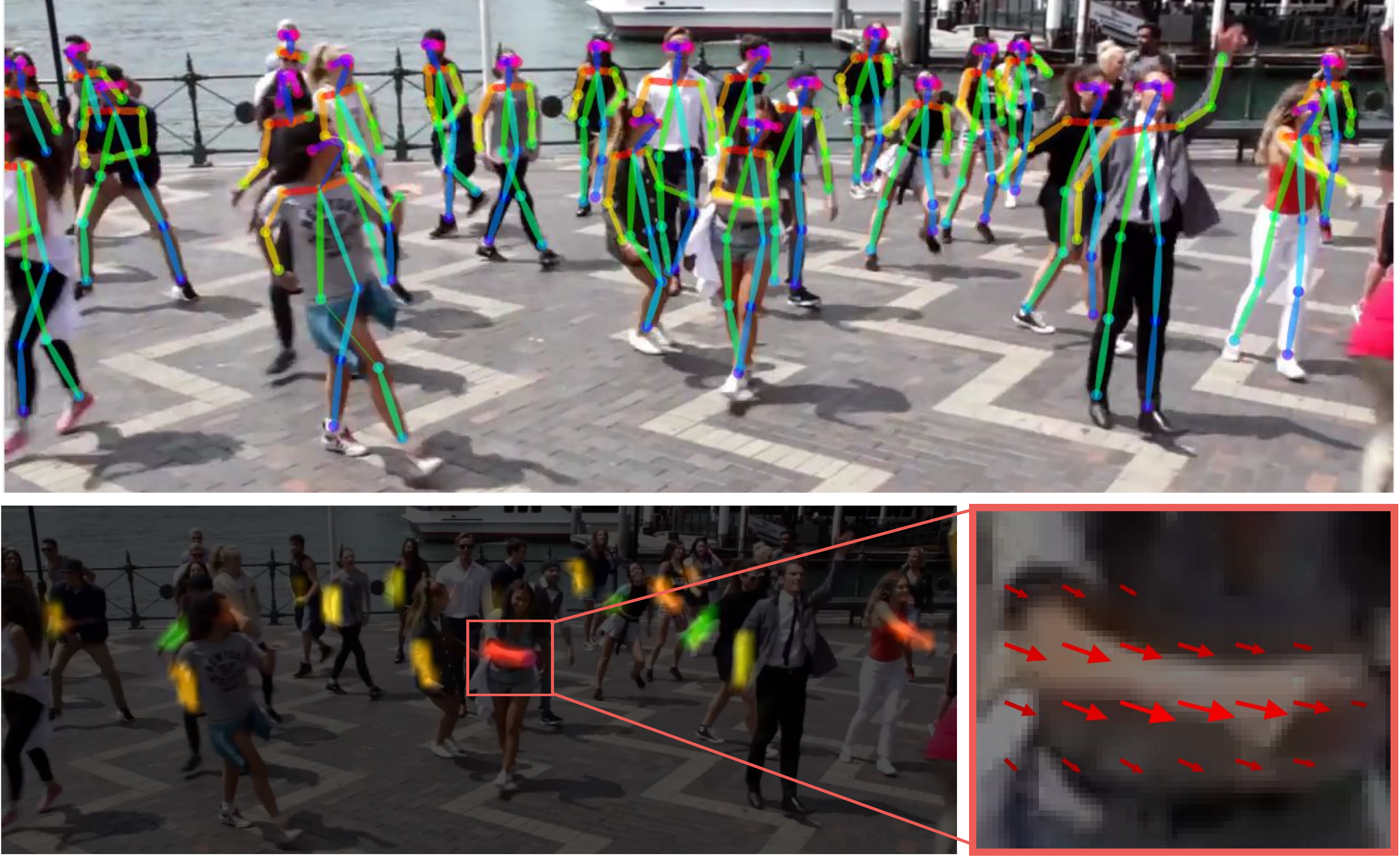} \\
    \caption{\textbf{Top:} Multi-person pose estimation. Body parts belonging to the same person are linked. \textbf{Bottom left:} Part Affinity Fields (PAFs) corresponding to the limb connecting right elbow and right wrist. The color encodes orientation. \textbf{Bottom right:} A zoomed in view of the predicted PAFs. At each pixel in the field, a 2D vector encodes the position and orientation of the limbs.}
    \vspace{-15pt}
    \label{fig:teaser}
\end{figure}

A common approach~\cite{pishchulin2012articulated,gkioxari2014using,sun2011articulated,iqbal2016multi,papandreou2017towards} is to employ a person detector and perform single-person pose estimation for each detection. These top-down approaches directly leverage existing techniques for single-person pose estimation \cite{newell2016stacked, wei2016convolutional, ouyang2014multi,tompson2014efficient,tompson2014joint,Chen_NIPS14,toshev2014deeppose,belagiannis2016recurrent,bulat2016human,pfister2015flowing}, but suffer from early commitment: if the person detector fails--as it is prone to do when people are in close proximity--there is no recourse to recovery. Furthermore, the runtime of these top-down approaches is proportional to the number of people: for each detection, a single-person pose estimator is run, and the more people there are, the greater the computational cost. In contrast, bottom-up approaches are attractive as they offer robustness to early commitment and have the potential to decouple runtime complexity from the number of people in the image. Yet, bottom-up approaches do not directly use global contextual cues from other body parts and other people. In practice, previous bottom-up methods \cite{pishchulin2015deepcut,insafutdinov2016deepercut} do not retain the gains in efficiency as the final parse requires costly global inference. For example, the seminal work of Pishchulin et al.~\cite{pishchulin2015deepcut} proposed a bottom-up approach that jointly labeled part detection candidates and associated them to individual people. However, solving the integer linear programming problem over a fully connected graph is an NP-hard problem and the average processing time is on the order of hours. Insafutdinov et al.~\cite{insafutdinov2016deepercut} built on~\cite{pishchulin2015deepcut} with stronger part detectors based on ResNet~\cite{he2015deep} and image-dependent pairwise scores, and
 vastly improved the runtime, but the method still takes several minutes per image, with a limit on the number of part proposals. The pairwise representations used in ~\cite{insafutdinov2016deepercut}, are difficult to regress precisely and thus a separate logistic regression is required.

\begin{figure*}[t]
    \centering
    \includegraphics[width=1\linewidth]{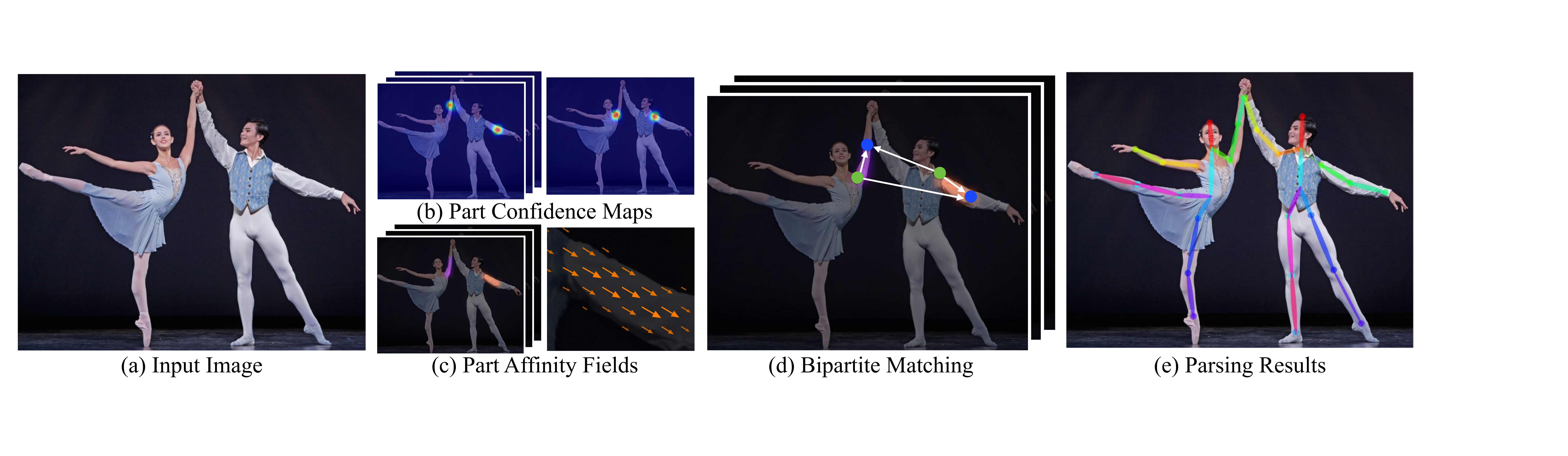} \\
 \vspace{-2pt}
    \caption{Overall pipeline. Our method takes the entire image as the input for a two-branch CNN to jointly predict confidence maps for body part detection, shown in (b), 
    and part affinity fields for parts association, shown in (c). 
    The parsing step performs a set of bipartite matchings to associate body parts candidates (d). We finally assemble them into full body poses for all people in the image (e).}
    \vspace{-12pt}
    \label{fig:pipeline}
\end{figure*}

In this paper, we present an efficient method for multi-person pose estimation with state-of-the-art accuracy on multiple public benchmarks. We present the first bottom-up representation of association scores via Part Affinity Fields (PAFs), a set of 2D vector fields that encode the location and orientation of limbs over the image domain. We demonstrate that simultaneously inferring these bottom-up representations of detection and association encode global context sufficiently well to allow a greedy parse to achieve high-quality results, at a fraction of the computational cost. We have publically released the \href{https://github.com/ZheC/Realtime_Multi-Person_Pose_Estimation}{code} for full reproducibility, presenting the first realtime system for multi-person 2D pose detection.

\section{Method}

Fig.~\ref{fig:pipeline} illustrates the overall pipeline of our method. The system takes, as input, a color image of size $w \times h$ (Fig.~\ref{fig:pipeline}a) and produces, as output, the 2D locations of anatomical keypoints for each person in the image (Fig.~\ref{fig:pipeline}e). First, a feedforward network simultaneously predicts a set of 2D confidence maps $\mathbf{S}$ of body part locations (Fig.~\ref{fig:pipeline}b) and a set of 2D vector fields $\mathbf{L}$ of part affinities, which encode the degree of association between parts (Fig.~\ref{fig:pipeline}c). The set $\mathbf{S}=(\mathbf{S}_1, \mathbf{S}_2, ..., \mathbf{S}_J)$ has $J$ confidence maps, one per part, where $\mathbf{S}_{j}\in \mathds{R}^{w\times h}$, $j \in \{1\ldots J\}$. The set $\mathbf{L} = (\mathbf{L}_1, \mathbf{L}_2, ..., \mathbf{L}_C)$ has $C$ vector fields, one per limb\footnote{We refer to part pairs as limbs for clarity, despite the fact that some pairs are not human limbs (e.g., the face).}, where $\mathbf{L}_{c}\in \mathds{R}^{w\times h \times 2}$, $c \in \{1\ldots C\}$, each image location in $\mathbf{L}_c$ encodes a 2D vector (as shown in Fig.~\ref{fig:teaser}). Finally, the confidence maps and the affinity fields are parsed by greedy inference (Fig.~\ref{fig:pipeline}d) to output the 2D keypoints for all people in the image.

\begin{figure}[t]
\begin{center}
\includegraphics[width=\linewidth]{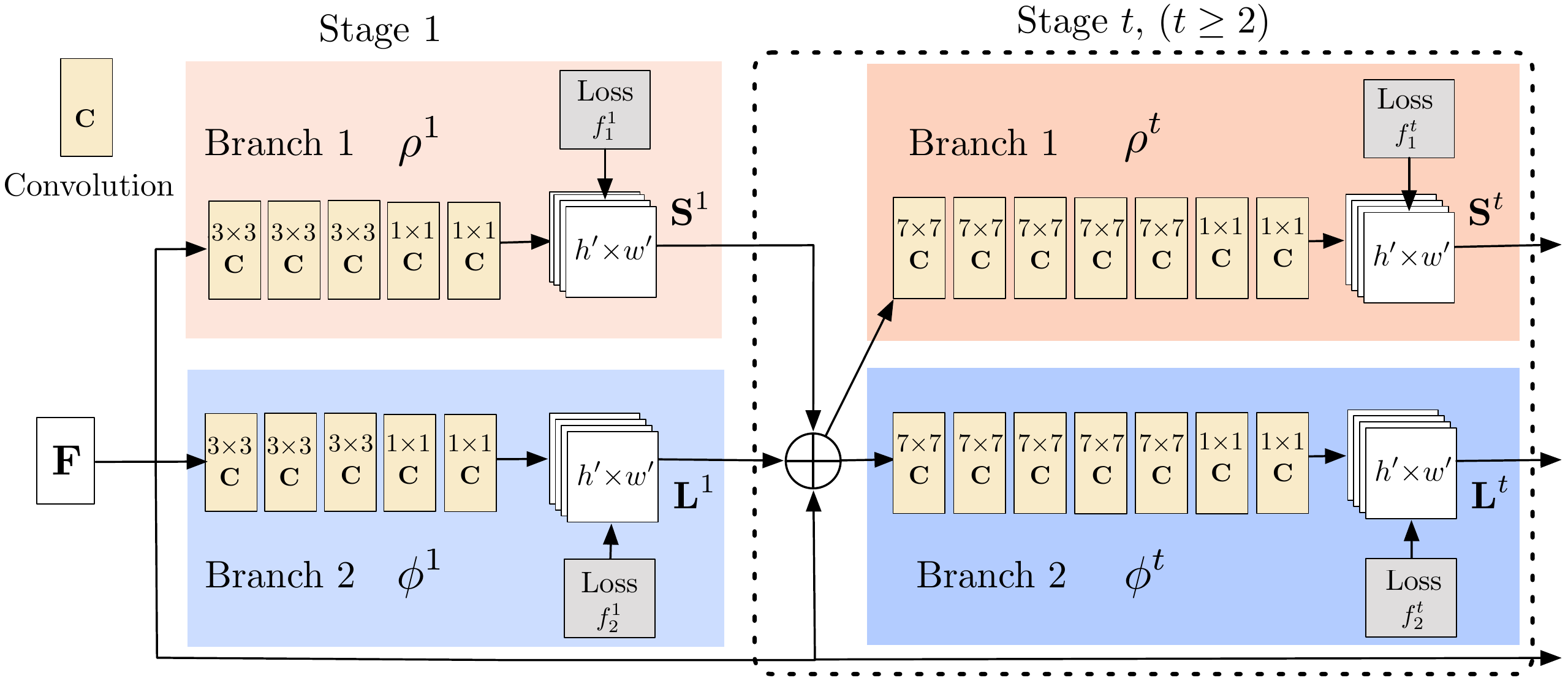}
\end{center}
\vspace{-8pt}
   \caption{Architecture of the two-branch multi-stage CNN. Each stage in the first branch predicts confidence maps $\mathbf{S}^t$, and each stage in the second branch predicts PAFs $\mathbf{L}^t$. After each stage, the predictions from the two branches, along with the image features, are concatenated for next stage.} 
   \vspace{-8pt}
\label{fig:arch}
\end{figure}
 
\subsection{Simultaneous Detection and Association}
\label{sec:arch}
Our architecture, shown in Fig.~\ref{fig:arch}, simultaneously predicts detection confidence maps and affinity fields that encode part-to-part association. The network is split into two branches: the top branch, shown in beige, predicts the confidence maps, and the bottom branch, shown in blue, predicts the affinity fields. Each branch is an iterative prediction architecture, following Wei et al.~\cite{wei2016convolutional}, which  refines the predictions over successive stages, $t\in \{1, \ldots, T\}$, with intermediate supervision at each stage.

The image is first analyzed by a convolutional network (initialized by the first 10 layers of VGG-19~\cite{simonyan2014very} and fine-tuned), generating a set of feature maps $\mathbf{F}$ that is input to the first stage of each branch. At the first stage, the network produces a set of detection confidence maps $\mathbf{S}^1 = \rho^{1}(\mathbf{F})$ and a set of part affinity fields $\mathbf{L}^1 = \phi^{1}(\mathbf{F})$, where $\rho^1$ and $\phi^1$ are the CNNs for inference at Stage 1. In each subsequent stage, the predictions from both branches in the previous stage, along with the original image features $\mathbf{F}$, are concatenated and used to produce refined predictions, 
\begin{eqnarray}
\mathbf{S}^t &=& \rho^{t}(\mathbf{F}, \mathbf{S}^{t-1}, \mathbf{L}^{t-1}),\; \forall t \geq 2,
\\
\mathbf{L}^t &=& \phi^{t}(\mathbf{F}, \mathbf{S}^{t-1}, \mathbf{L}^{t-1}),\; \forall t \geq 2, 
\end{eqnarray}
where $\rho^t$ and $\phi^t$ are the CNNs for inference at Stage $t$.

 \begin{figure}[t!]
  \centering
  \includegraphics[width=1\linewidth]{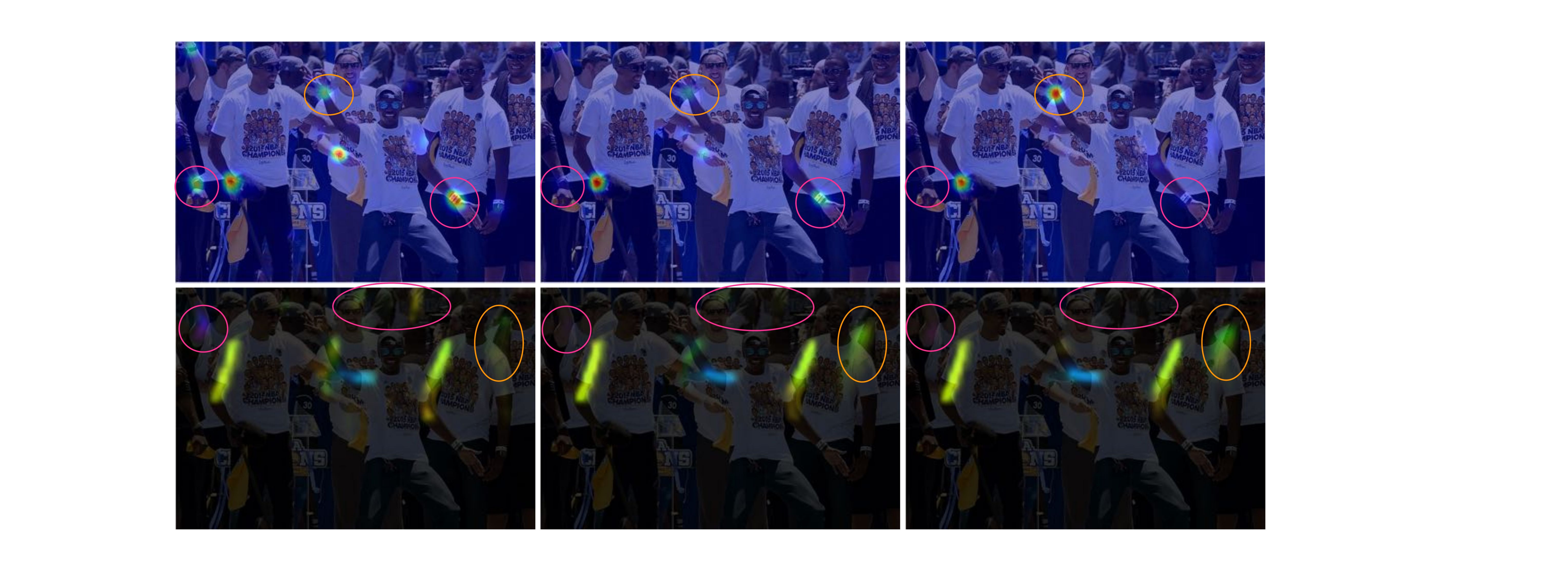}
  \\
  \vspace{-5pt}
  {\footnotesize Stage 1}\hspace{50pt} {\footnotesize Stage 3} \hspace{50pt} {\footnotesize Stage 6}\\
  \vspace{2pt}
  \caption{\label{fig8} Confidence maps of the right wrist (first row) and PAFs (second row) of right forearm across stages. Although there is confusion between left and right body parts and limbs in early stages, the estimates are increasingly refined through global inference in later stages, as shown in the highlighted areas.}
  \vspace{-5pt}
  \label{fig:stages}
\end{figure}
 
Fig.~\ref{fig8} shows the refinement of the confidence maps and affinity fields across stages. To guide the network to iteratively predict confidence maps of body parts in the first branch and PAFs in the second branch, we apply two loss functions at the end of each stage, one at each branch respectively. We use an $L_2$ loss between the estimated predictions and the groundtruth maps and fields. Here, we weight the loss functions spatially to address a practical issue that some datasets do not completely label all people. Specifically, the loss functions at both branches at stage $t$ are:
 \begin{eqnarray}
     f^t_\mathbf{S} &=& \sum_{j = 1}^{J} \sum_{\mathbf{p}}  \mathbf{W(\mathbf{p})}\cdot \| \mathbf{S}^t_{j}(\mathbf{p})  -  \mathbf{S}_{j}^{*}(\mathbf{p}) \|^{2}_{2},\\
     f^t_\mathbf{L} &=& \sum_{c = 1}^{C} \sum_{\mathbf{p}}  \mathbf{W(\mathbf{p})}\cdot \| \mathbf{L}^t_{c}(\mathbf{p})  -  \mathbf{L}_{c}^{*}(\mathbf{p}) \|^{2}_{2},
    \label{eqn:localobjective2}
\end{eqnarray}
where $\mathbf{S}_{j}^{*}$ is the groundtruth part confidence map, $\mathbf{L}_{c}^{*}$ is the groundtruth part affinity vector field, $\mathbf{W}$ is a binary mask with $\mathbf{W}(\mathbf{p}) = 0$ when the annotation is missing at an image location $\mathbf{p}$. The mask is used to avoid penalizing the true positive predictions during training. 
The intermediate supervision at each stage addresses the vanishing gradient problem by replenishing the gradient periodically~\cite{wei2016convolutional}. The overall objective is
\begin{equation}\label{eq:overall}
    f=\sum_{t=1}^{T} (f_\mathbf{S}^t + f_\mathbf{L}^t).
\end{equation}

\subsection{Confidence Maps for Part Detection}
\label{sec:confidence}
To evaluate $f_\mathbf{S}$ in Eq.~\eqref{eq:overall} during training, we generate the groundtruth confidence maps $\mathbf{S}^*$ from the annotated 2D keypoints. Each confidence map is a 2D representation of the belief that a particular body part occurs at each pixel location. Ideally, if a single person occurs in the image, a single peak should exist in each confidence map if the corresponding part is visible; if multiple people occur, there should be a peak corresponding to each visible part $j$ for each person $k$. 

We first generate individual confidence maps $\mathbf{S}_{j,k}^*$ for each person $k$. Let $\mathbf{x}_{j,k}\in\mathds{R}^2$ be the groundtruth position of body part $j$ for person $k$ in the image. The value at location $\mathbf{p}\in\mathds{R}^2$ in $\mathbf{S}_{j,k}^*$ is defined as, 
\begin{equation}
\mathbf{S}_{j,k}^{*}(\mathbf{p}) = \operatorname{exp}\left(-\frac{||\mathbf{p}-\mathbf{x}_{j,k}||_2^2}{\sigma^2}\right),
\end{equation}
where $\sigma$ controls the spread of the peak. The groundtruth confidence map to be predicted by the network is an aggregation of the individual confidence maps via a max operator,
\begin{equation}
\mathbf{S}_j^{*}(\mathbf{p}) = \underset{k}{\operatorname{max}}\, \mathbf{S}_{j,k}^{*}(\mathbf{p}).
\end{equation}

\begin{wrapfigure}{r}{0.43\columnwidth}
\vspace{-0.15in}
\includegraphics[width=1\linewidth]{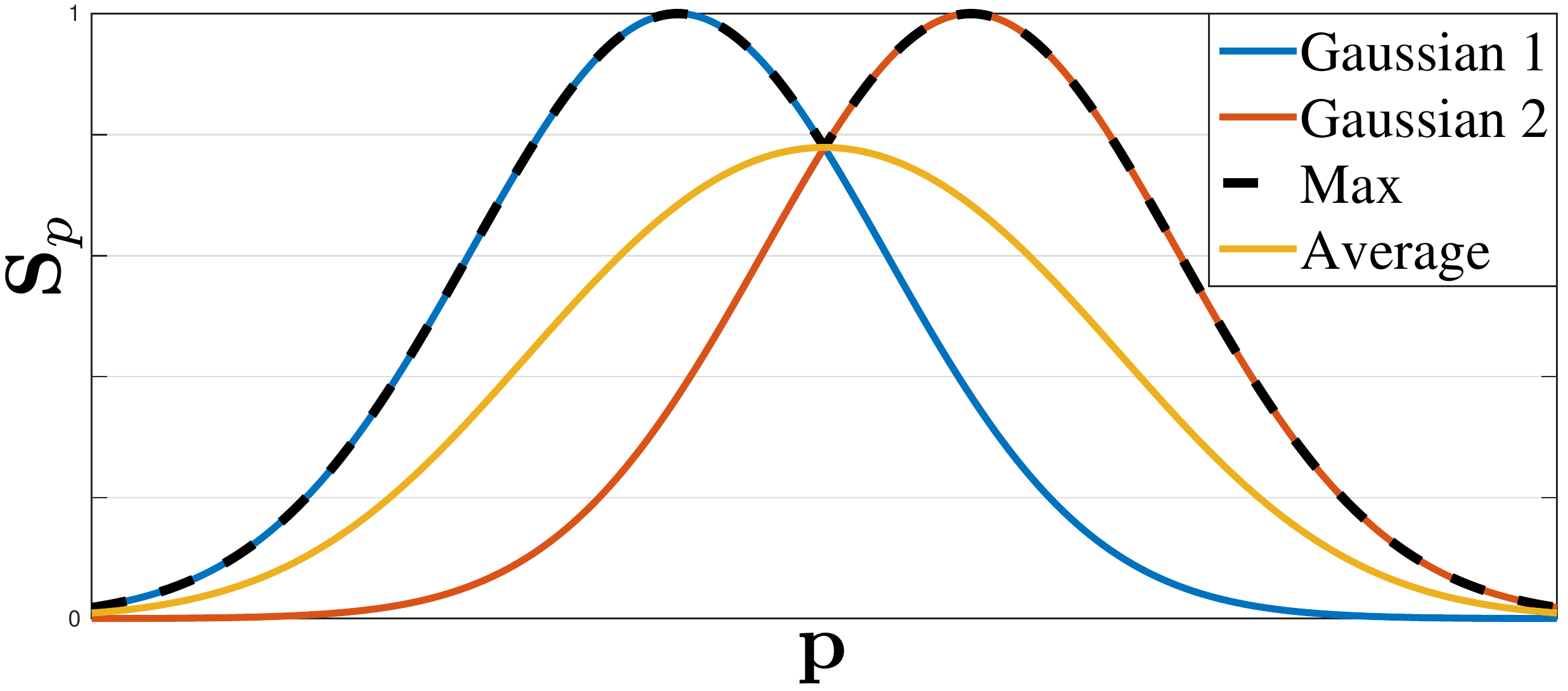}
\vspace{-0.3in}
\end{wrapfigure}
We take the maximum of the confidence maps instead of the average so that the precision of close by peaks remains distinct, as illustrated in the right figure. At test time, we predict confidence maps (as shown in the first row of Fig.~\ref{fig8}), and obtain body part candidates by performing non-maximum suppression.

\subsection{Part Affinity Fields for Part Association}

\begin{figure}[t!]
    \centering
    \includegraphics[width=1\linewidth]{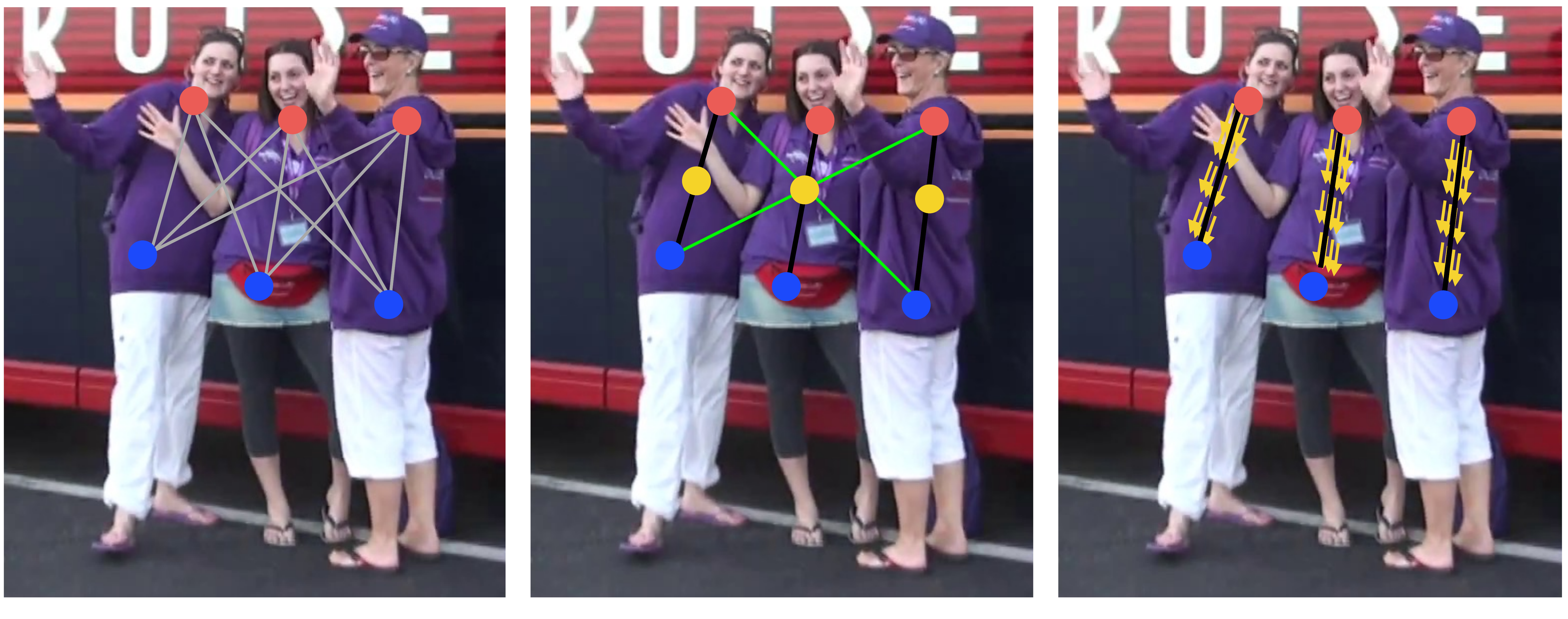}\\
    \vspace{-6pt}
  {\footnotesize (a)}\hspace{65pt}  {\footnotesize (b) }\hspace{65pt} {\footnotesize (c)}\\
    \caption{Part association strategies. (a) The body part detection candidates (red and blue dots) for two body part types and all connection candidates (grey lines). (b) The connection results using the midpoint (yellow dots) representation: correct connections (black lines) and incorrect connections (green lines) that also satisfy the incidence constraint. (c) The results using PAFs (yellow arrows). By encoding position and orientation over the support of the limb, PAFs eliminate false associations.}
    \vspace{-10pt}
    \label{fig:ambiguity}
\end{figure}

Given a set of detected body parts (shown as the red and blue points in Fig.~\ref{fig:ambiguity}a), how do we assemble them to form the full-body poses of an unknown number of people? We need a confidence measure of the association for each pair of body part detections, i.e., that they belong to the same person. One possible way to measure the association is to detect an additional midpoint between each pair of parts on a limb, and check for its incidence between candidate part detections, as shown in Fig.~\ref{fig:ambiguity}b. However, when people crowd together---as they are prone to do---these midpoints are likely to support false associations (shown as green lines in Fig.~\ref{fig:ambiguity}b). Such false associations arise due to two limitations in the representation: (1) it encodes only the position, and not the orientation, of each limb; (2) it reduces the region of support of a limb to a single point.

To address these limitations, we present a novel feature representation called {\em part affinity fields} that preserves both location and orientation information across the region of support of the limb (as shown in Fig.~\ref{fig:ambiguity}c). The part affinity is a 2D vector field for each limb, also shown in Fig.~\ref{fig:teaser}d: for each pixel in the area belonging to a particular limb, a 2D vector encodes the direction that points from one part of the limb to the other. Each type of limb has a corresponding affinity field joining its two associated body parts. 

Consider a single limb shown in the figure below. Let $\mathbf{x}_{j_1,k}$ and $\mathbf{x}_{j_2,k}$ be the groundtruth positions of body parts $j_1$ and $j_2$ from the limb $c$ for person $k$ in the image. If a point  
\begin{wrapfigure}{r}{0.42\columnwidth}
\vspace{-0.15in}
\includegraphics[width=1\linewidth]{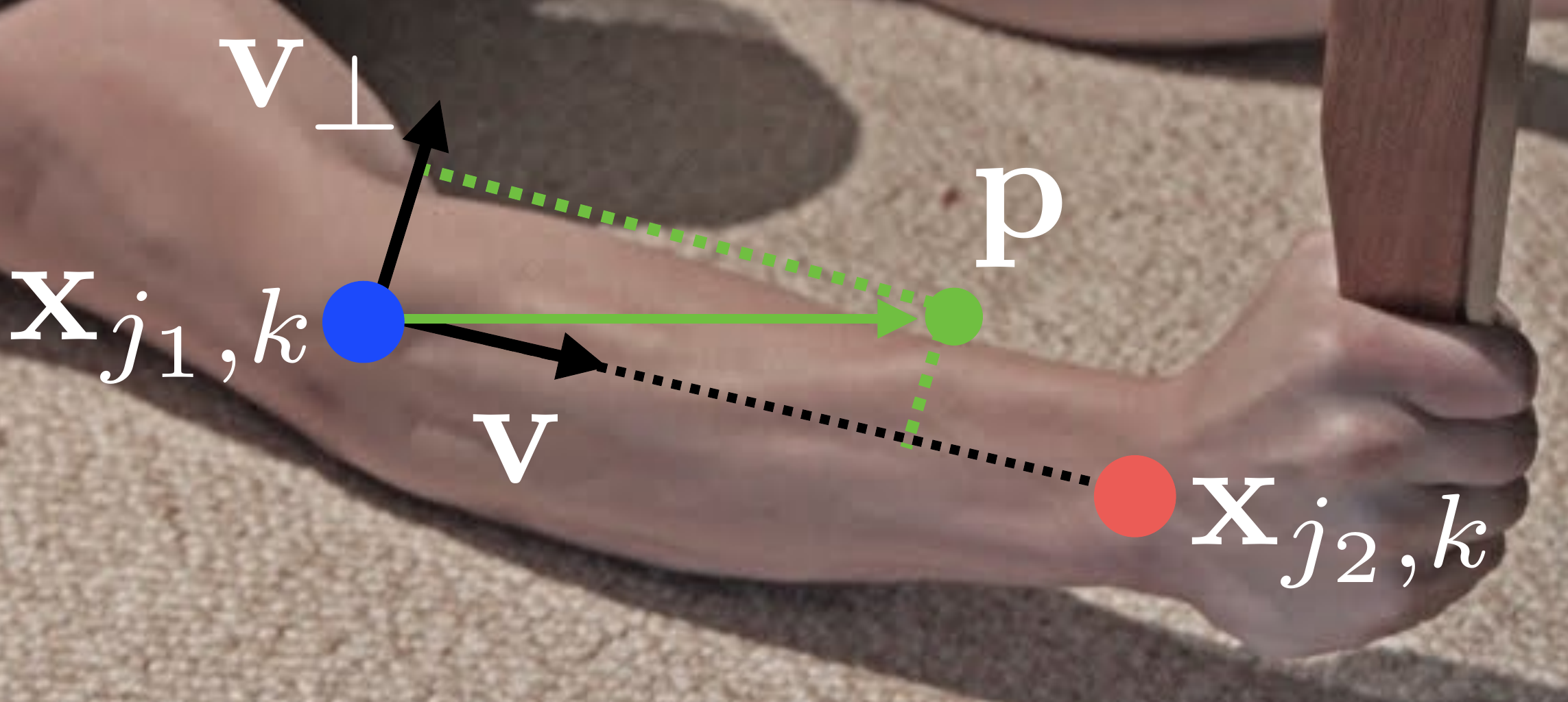}
\vspace{-0.3in}
\end{wrapfigure}
$\mathbf{p}$ lies on the limb, the value at $\mathbf{L}_{c,k}^*(\mathbf{p})$ is a unit vector that points from $j_1$ to $j_2$; for all other points, the vector is zero-valued. 

To evaluate $f_\mathbf{L}$ in Eq.~\ref{eq:overall} during training, we define the groundtruth part affinity vector field, $\mathbf{L}_{c,k}^{*}$, at an image point $\mathbf{p}$ as
\begin{equation}
\mathbf{L}_{c,k}^{*}(\mathbf{p}) = \begin{cases}
\mathbf{v} &\textrm{if\ }\mathbf{p}\ \textrm{on limb } c,k\\
\mathbf{0} &\textrm{otherwise.\ }
\end{cases}
\end{equation}
Here, $\mathbf{v}=({\mathbf{x}_{j_2,k}-\mathbf{x}_{j_1,k}})/||\mathbf{x}_{j_2,k}-\mathbf{x}_{j_1,k}||_2$ is the unit vector in the direction of the limb. The set of points on the limb is defined as those within a distance threshold of the line segment, i.e., those points $\mathbf{p}$ for which
\[
0\,{\leq}\,{\mathbf{v}\cdot(\mathbf{p}-\mathbf{x}_{j_1,k})}\,{\leq}\,l_{c,k}\ \textrm{ and }\ |\mathbf{v}_\perp \cdot(\mathbf{p}-\mathbf{x}_{j_1,k})|\,{\leq}\,\sigma_l,
\]
where the limb width $\sigma_l$ is a distance in pixels, the limb length is  $l_{c,k} = ||\mathbf{x}_{j_2,k}-\mathbf{x}_{j_1,k}||_2$, and $\mathbf{v}_\perp$ is a vector perpendicular to $\mathbf{v}$. 

The groundtruth part affinity field averages the affinity fields of all people in the image,
\begin{equation}
\mathbf{L}_{c}^{*}(\mathbf{p}) = \frac{1}{n_c(\mathbf{p})}\sum_{k}\mathbf{L}_{c,k}^{*}(\mathbf{p}),  \vspace{-6pt}
\end{equation}
where $n_c(\mathbf{p})$ is the number of non-zero vectors at point $\mathbf{p}$ across all $k$ people (i.e., the average at pixels where limbs of different people overlap).

During testing, we measure association between candidate part detections by computing the line integral over the corresponding PAF, along the line segment connecting the candidate part locations. In other words, we measure the alignment of the predicted PAF with the candidate limb that would be formed by connecting the detected body parts. Specifically, for two candidate part locations $\mathbf{d}_{j_1}$ and $\mathbf{d}_{j_2}$, we sample the predicted part affinity field, $\mathbf{L}_{c}$ along the line segment to measure the confidence in their association:
\begin{equation}
E = \int_{u=0}^{u=1}  \mathbf{L}_{c}\left( \mathbf{p}(u) \right) \cdot \frac{\mathbf{d}_{j_2} - \mathbf{d}_{j_1}}{||\mathbf{d}_{j_2} - \mathbf{d}_{j_1}||_2}du,
\label{eq:score1}
\end{equation}
where $\mathbf{p}(u)$ interpolates the position of the two body parts $d_{j_1}$ and $d_{j_2}$,
\begin{equation}
 \mathbf{p}(u) = (1-u) \mathbf{d}_{j_1} + u \mathbf{d}_{j_2}.
\end{equation}
\noindent In practice, we approximate the integral by sampling and summing uniformly-spaced values of $u$.

\begin{figure}[t!]
    \centering
    \includegraphics[width=1\linewidth]{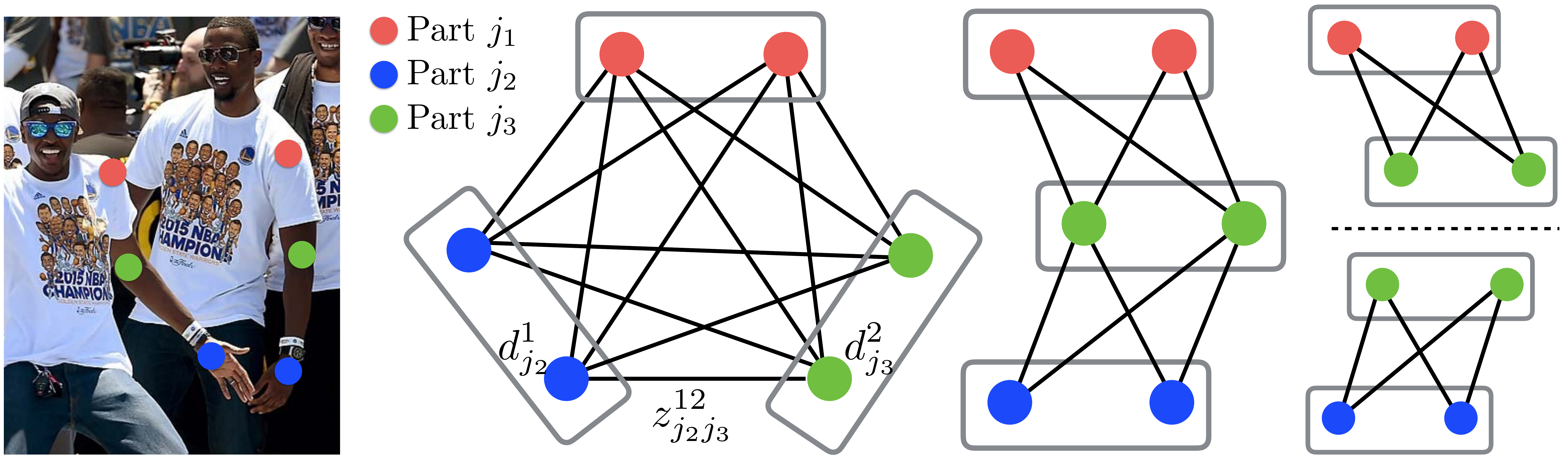}\\
    \vspace{-6pt}
   \hspace{0.2cm} {\footnotesize (a)}\hspace{2.3cm} {\footnotesize (b)}\hspace{1.8cm}  {\footnotesize (c) }\hspace{1.1cm} {\footnotesize (d)}\\
    \caption{Graph matching. (a) Original image with part detections (b) $K$-partite graph (c) Tree structure (d) A set of bipartite graphs}
    \vspace{-15pt}
    \label{fig:GM}
\end{figure}

\subsection{Multi-Person Parsing using PAFs} 
We perform non-maximum suppression on the detection confidence maps to obtain a discrete set of part candidate locations. For each part, we may have several candidates, due to multiple people in the image or false positives (shown in Fig.~\ref{fig:GM}b). These part candidates define a large set of possible limbs. We score each candidate limb using the line integral computation on the PAF, defined in Eq.~\ref{eq:score1}. The problem of finding the optimal parse corresponds to a $K$-dimensional matching problem that is known to be NP-Hard~\cite{west2001introduction} (shown in Fig.~\ref{fig:GM}c). In this paper, we present a greedy relaxation that consistently produces high-quality matches. We speculate the reason is that the pair-wise association scores implicitly encode global context, due to the large receptive field of the PAF network.

Formally, we first obtain a set of body part detection candidates $\mathcal{D}_\mathcal{J}$ for multiple people, where $\mathcal{D}_\mathcal{J} = \{ \mathbf{d}_j^m : \text{for}\ {j \in \{1 \ldots J\}, m \in \{1 \ldots N_j\}\}}$, with $N_j$ the number of candidates of part $j$, and $\mathbf{d}_j^m\in\mathds{R}^2$ is the location of the $m$-th detection candidate of body part $j$. These part detection candidates still need to be associated with other parts from the same person---in other words, we need to find the pairs of part detections that are in fact connected limbs. We define a variable $z_{{j_1}{j_2}}^{mn} \in \{0,1\}$ to indicate whether two detection candidates $\mathbf{d}^m_{j_1}$ and $\mathbf{d}^n_{j_2}$ are connected, and the goal is to find the optimal assignment for the set of all possible connections, $\mathcal{Z} = \{ z_{{j_1}{j_2}}^{mn}: \text{for}\ {j_1,j_2 \in \{1 \ldots J\}, m \in \{1 \ldots N_{j_1}\}, n \in \{1 \ldots N_{j_2}\}\}}$.

If we consider a {\em single} pair of parts $j_1$ and $j_2$ (e.g., neck and right hip) for the $c$-th limb, finding the optimal association reduces to a maximum weight bipartite graph matching problem~\cite{west2001introduction}. This case is shown in Fig.~\ref{fig:ambiguity}b. In this graph matching problem, nodes of the graph are the body part detection candidates $\mathcal{D}_{j_1}$ and $\mathcal{D}_{j_2}$, and the edges are all possible connections between pairs of detection candidates. Additionally, each edge is weighted by Eq.~\ref{eq:score1}---the part affinity aggregate. A matching in a bipartite graph is a subset of the edges chosen in such a way that no two edges share a node. Our goal is to find a matching with maximum weight for the chosen edges, 
\vspace{-5pt}
\begin{eqnarray}
\label{eqn:opt4}
\max_{\mathcal{Z}_c} E_c = \max_{\mathcal{Z}_c} \sum_{m \in \mathcal{D}_{j_1}} \sum_{n \in \mathcal{D}_{j_2} } E_{mn} \cdot z_{{j_1}{j_2}}^{mn},
\end{eqnarray}
\vspace{-10pt}
\begin{eqnarray}
\label{eqn:opt5}
\mathrm{s.t.}&~&\forall {m \in \mathcal{D}_{j_1}}, \sum_{n \in \mathcal{D}_{j_2}} z_{{j_1}{j_2}}^{mn} \leq 1,
\\ \label{eqn:opt6}
&~&\forall {n \in \mathcal{D}_{j_2}}, \sum_{m \in \mathcal{D}_{j_1}} z_{{j_1}{j_2}}^{mn} \leq 1,
\end{eqnarray}
\vspace{-15pt}

\noindent where $E_c$ is the overall weight of the matching from limb type $c$, $\mathcal{Z}_c$ is the subset of $\mathcal{Z}$ for limb type $c$, $E_{mn}$ is the part affinity between parts $\mathbf{d}^m_{j_1}$ and $\mathbf{d}^n_{j_2}$ defined in Eq.~\ref{eq:score1}.
Eqs.~\ref{eqn:opt5} and~\ref{eqn:opt6} enforce no two edges share a node, i.e., no two limbs of the same type (e.g., left forearm) share a part. We can use the Hungarian algorithm~\cite{kuhn1955hungarian} to obtain the optimal matching. 

When it comes to finding the full body pose of multiple people, determining $\mathcal{Z}$ is a $K$-dimensional matching problem. This problem is NP Hard~\cite{west2001introduction} and many relaxations exist. In this work, we add two relaxations to the optimization, specialized to our domain. First, we choose a minimal number of edges to obtain a spanning tree skeleton of human pose rather than using the complete graph, as shown in Fig.~\ref{fig:GM}c. Second, we further decompose the matching problem into a set of bipartite matching subproblems and determine the matching in adjacent tree nodes independently, as shown in Fig.~\ref{fig:GM}d. We show detailed comparison results in Section~\ref{sec:mpii}, which demonstrate that minimal greedy inference well-approximate the global solution at a fraction of the computational cost. The reason is that the relationship between adjacent tree nodes is modeled explicitly by PAFs, but internally, the relationship between nonadjacent tree nodes is implicitly modeled by the CNN. This property emerges because the CNN is trained with a large receptive field, and PAFs from non-adjacent tree nodes also influence the predicted PAF. 

With these two relaxations, the optimization is decomposed simply as:
\vspace{-5pt}
\begin{equation}
\max_{\mathcal{Z}} E = \sum_{c = 1}^{C} \max_{\mathcal{Z}_c} E_c.
\label{eqn:opt1}
\end{equation}
We therefore obtain the limb connection candidates for each limb type independently using Eqns.~\ref{eqn:opt4}-~\ref{eqn:opt6}. With all limb connection candidates, we can assemble the connections that share the same part detection candidates into full-body poses of multiple people. Our optimization scheme over the tree structure is orders of magnitude faster than the optimization over the fully connected graph~\cite{pishchulin2015deepcut,insafutdinov2016deepercut}.  

\section{Results}

We evaluate our method on two benchmarks for multi-person pose estimation: (1) the MPII human multi-person dataset~\cite{andriluka20142d} and (2) the COCO 2016 keypoints challenge dataset~\cite{lin2014microsoft}. These two datasets collect images in diverse scenarios that contain many real-world challenges such as crowding, scale variation, occlusion, and contact. Our approach set the state-of-the-art on the inaugural COCO 2016 keypoints challenge~\cite{COCOkeypoint}, and significantly exceeds the previous state-of-the-art result on the MPII multi-person benchmark. We also provide runtime analysis to quantify the efficiency of the system. Fig.~\ref{fig:qua} shows some qualitative results from our algorithm. 

%

%

\begin{table}[t]
\begin{center}
\resizebox{0.485\textwidth}{!}{
\begin{tabular}{l | c c c c c c c |c |c}
\hline
Method & Hea & Sho & Elb & Wri & Hip & Kne & Ank & \textbf{mAP} & s/image \\
\hline
\multicolumn{10}{ c }{Subset of 288 images as in \cite{pishchulin2015deepcut}} \\
Deepcut~\cite{pishchulin2015deepcut} &73.4 &71.8 &57.9 &39.9 &56.7 &44.0 &32.0 &54.1 &57995\\
Iqbal et al.~\cite{iqbal2016multi} & 70.0 &65.2 &56.4 &46.1 &52.7 &47.9 &44.5 &54.7 & 10\\
DeeperCut~\cite{insafutdinov2016deepercut} &87.9 &84.0 &71.9 &63.9 &68.8 &63.8 &58.1 &71.2 &230\\
Ours & \textbf{93.7} & \textbf{91.4} & \textbf{81.4} & \textbf{72.5} & \textbf{77.7} & \textbf{73.0} & \textbf{68.1} & \textbf{79.7} & \textbf{0.005}\\
\hline
\multicolumn{10}{ c }{Full testing set} \\
DeeperCut~\cite{insafutdinov2016deepercut} &78.4 &72.5 &60.2 &51.0 &57.2 &52.0 &45.4 &59.5 & 485\\
Iqbal et al.~\cite{iqbal2016multi} & 58.4 & 53.9 &44.5 &35.0	&42.2 &36.7	&31.1 &43.1 & 10\\
Ours (one scale) & 89.0  & 84.9  & 74.9  & 64.2  & 71.0  & 65.6 & 58.1 & 72.5 & 0.005\\
Ours  & \textbf{91.2}  & \textbf{87.6}  & \textbf{77.7}  & \textbf{66.8} & \textbf{75.4}  & \textbf{68.9} & \textbf{61.7} & \textbf{75.6} & \textbf{0.005}\\
\hline
\end{tabular}}
\end{center}
\vspace{-5pt}
\caption{Results on the MPII dataset. Top: Comparison result on the testing subset. Middle: Comparison results on the whole testing set. Testing without scale search is denoted as ``(one scale)".}
\vspace{-5pt}
\label{table:mpi}
\end{table}

\begin{table}[t]
\begin{center}
\resizebox{0.485\textwidth}{!}{
\begin{tabular}{l | c c c c c c c |c |c}
\hline
Method & Hea & Sho & Elb & Wri & Hip & Kne & Ank & \textbf{mAP} & s/image \\
\hline
Fig.~\ref{fig:GM}b &91.8 &\textbf{90.8} &80.6 &69.5 &78.9 &71.4 &63.8 &78.3 & 362\\ 
Fig.~\ref{fig:GM}c &92.2 &90.8 &80.2 &69.2 &78.5 &70.7 &62.6 &77.6 & 43\\
Fig.~\ref{fig:GM}d &92.0 &90.7 &80.0 &69.4 &78.4 &70.1 &62.3 &77.4 & 0.005\\
Fig.~\ref{fig:GM}d (sep) &\textbf{92.4}  & 90.4  & \textbf{80.9}  & \textbf{70.8} & \textbf{79.5}  & \textbf{73.1} & \textbf{66.5} & \textbf{79.1} & \textbf{0.005} \\
\hline
\end{tabular}}
\end{center}
\vspace{-5pt}
\caption{Comparison of different structures on our validation set.}
\vspace{-15pt}
\label{table:fig5}
\end{table}

\subsection{Results on the MPII Multi-Person Dataset}\label{sec:mpii}

For comparison on the MPII dataset, we use the toolkit~\cite{pishchulin2015deepcut} to measure mean Average Precision (mAP) of all body parts based on the PCKh threshold. 
Table~\ref{table:mpi} compares mAP performance between our method and other approaches on the same subset of $288$ testing images as in~\cite{pishchulin2015deepcut}, and the entire MPI testing set, and self-comparison on our own validation set. Besides these measures, we compare the average inference/optimization time per image in seconds. For the $288$ images subset, our method outperforms previous state-of-the-art bottom-up methods~\cite{insafutdinov2016deepercut} by $8.5\%$ mAP. Remarkably, our inference time is $6$ orders of magnitude less. We report a more detailed runtime analysis in Section~\ref{sec:runtime}. For the entire MPII testing set, our method without scale search already outperforms previous state-of-the-art methods by a large margin, i.e., $13\%$ absolute increase on mAP. Using a 3 scale search ($\times0.7$, $\times1$ and $\times1.3$) further increases the performance to $75.6\%$ mAP. The mAP comparison with previous bottom-up approaches indicate the effectiveness of our novel feature representation, PAFs, to associate body parts. Based on the tree structure, our greedy parsing method achieves better accuracy than a graphcut optimization formula based on a fully connected graph structure~\cite{pishchulin2015deepcut,insafutdinov2016deepercut}. 

In Table~\ref{table:fig5}, we show comparison results on different skeleton structures as shown in Fig.~\ref{fig:GM} on our validation set, i.e., 343 images excluded from the MPII training set. We train our model based on a fully connected graph, and compare results by selecting all edges (Fig.~\ref{fig:GM}b, approximately solved by Integer Linear Programming), and minimal tree edges (Fig.~\ref{fig:GM}c, approximately solved by Integer Linear Programming, and Fig.~\ref{fig:GM}d, solved by the greedy algorithm presented in this paper). Their similar performance shows that it suffices to use minimal edges. We trained another model that only learns the minimal edges to fully utilize the network capacity---the method presented in this paper---that is denoted as Fig.~\ref{fig:GM}d (sep). This approach outperforms Fig.~\ref{fig:GM}c and even Fig.~\ref{fig:GM}b, while maintaining efficiency. The reason is that the much smaller number of part association channels (13 edges of a tree vs 91 edges of a graph) makes it easier for training convergence.

\begin{figure}[t]
\centering
 \vspace{5pt}
  \includegraphics[width=0.516\linewidth]{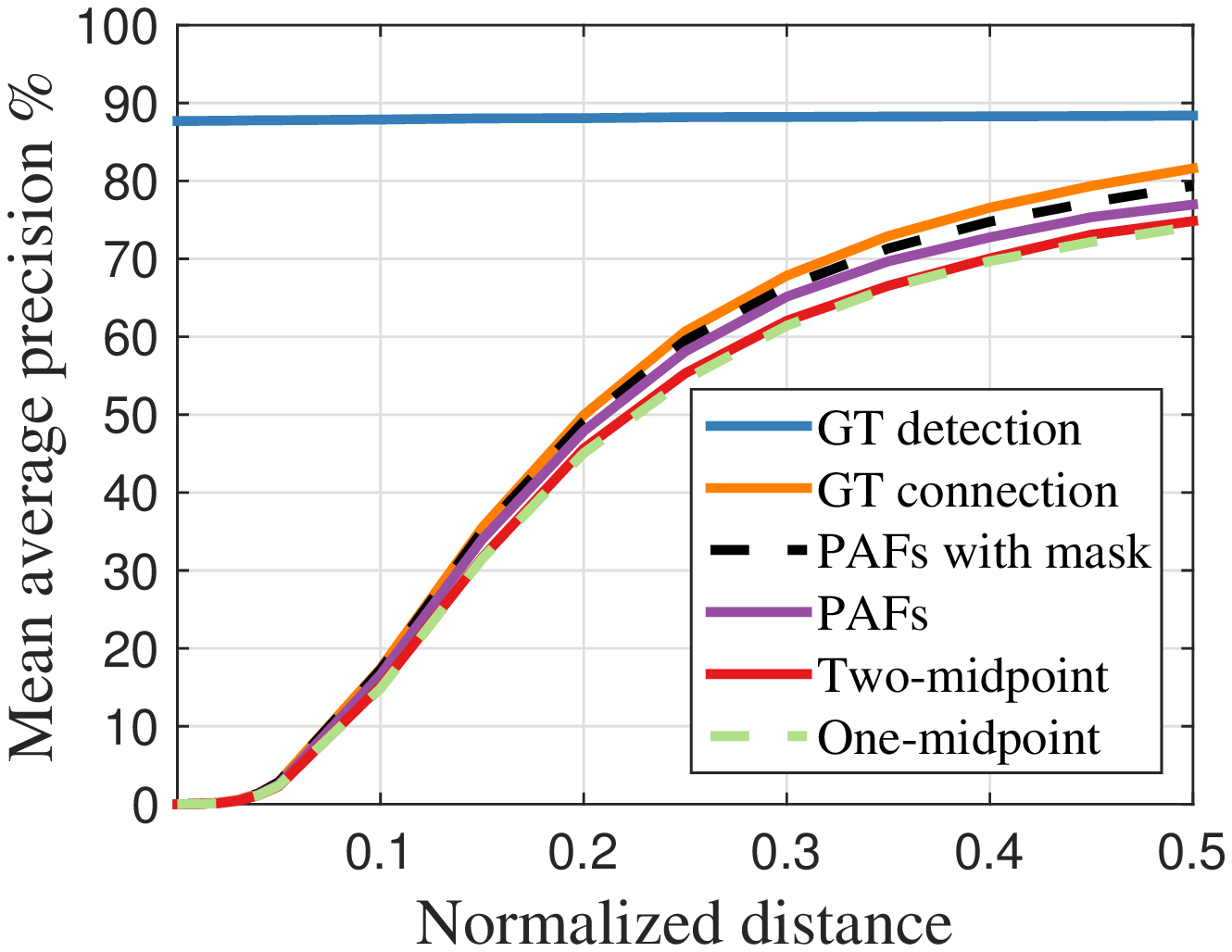}
    \hfill
  \includegraphics[width=0.4725\linewidth]{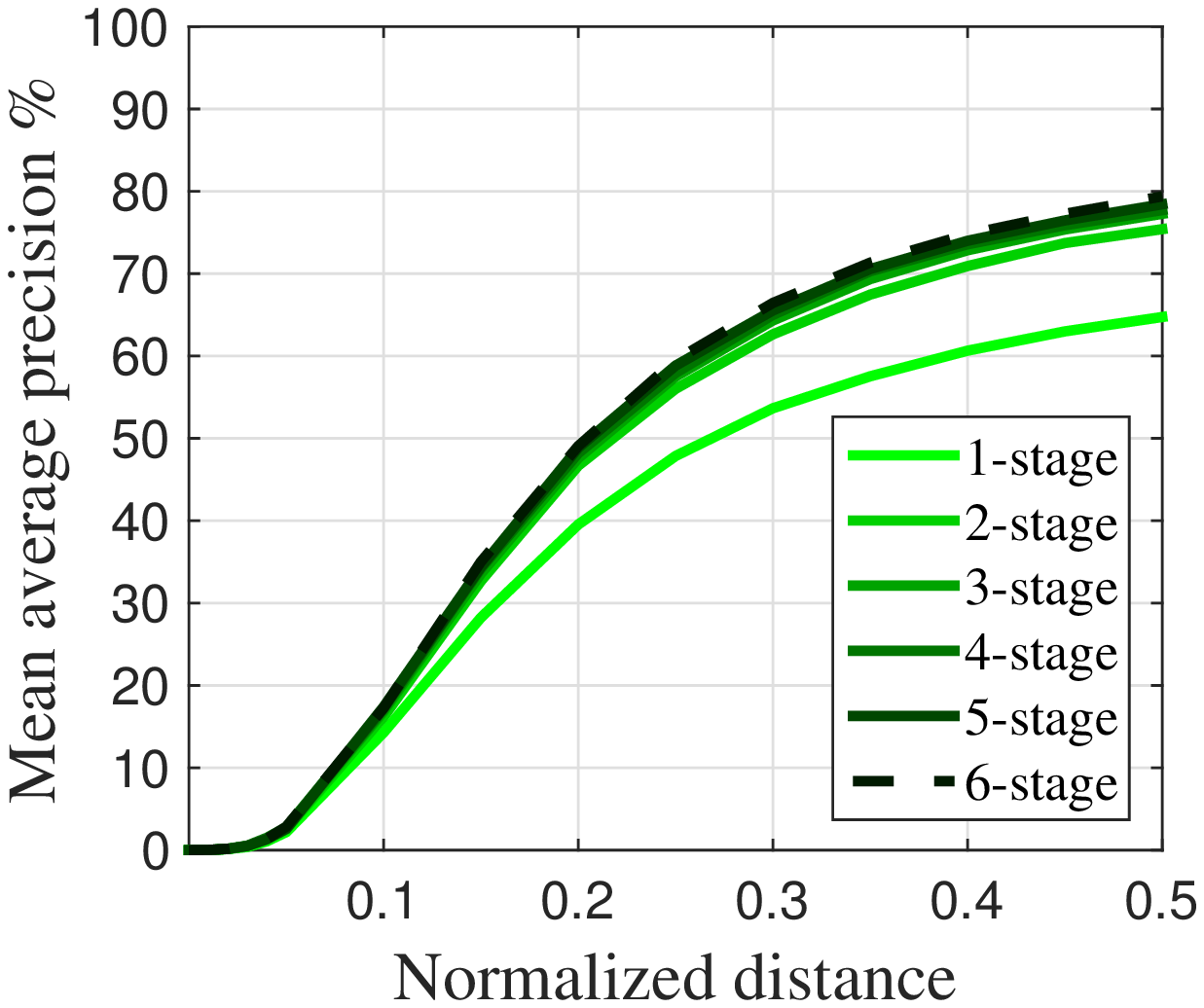}\\
    \vspace{-2pt}
  \hspace{15pt}{\footnotesize (a)}\hspace{110pt}  {\footnotesize (b) }\\
  \caption{mAP curves over different PCKh threshold on MPII validation set. (a) mAP curves of self-comparison experiments. 
  (b) mAP curves of PAFs across stages.}
  \vspace{-10pt}
\label{fig:curve}
 \end{figure}

Fig.~\ref{fig:curve}a shows an ablation analysis on our validation set. For the threshold of PCKh-0.5, the result using PAFs outperforms the results using the midpoint representation, specifically, it is $2.9\%$ higher than one-midpoint and $2.3\%$ higher than two intermediate points. The PAFs, which encodes both position and orientation information of human limbs, is better able to distinguish the common cross-over cases, e.g., overlapping arms. Training with masks of unlabeled persons further improves the performance by $2.3\%$ because it avoids penalizing the true positive prediction in the loss during training. If we use the ground-truth keypoint location with our parsing algorithm, we can obtain a mAP of $88.3\%$. In Fig.~\ref{fig:curve}a, the mAP of our parsing with GT detection is constant across different PCKh thresholds due to no localization error. Using GT connection with our keypoint detection achieves a mAP of $81.6\%$. It is notable that our parsing algorithm based on PAFs achieves a similar mAP as using GT connections ($79.4\%$ vs $81.6\%$). This indicates parsing based on PAFs is quite robust in associating correct part detections. 
Fig.~\ref{fig:curve}b shows a comparison of performance across stages. The mAP increases monotonically with the iterative refinement framework. Fig.~\ref{fig:stages} shows the qualitative improvement of the predictions over stages. 


\subsection{Results on the COCO Keypoints Challenge}
\label{sec:coco}
The COCO training set consists of over 100K person instances labeled with over 1 million total keypoints (i.e. body parts). The testing set contains ``test-challenge'', ``test-dev'' and ``test-standard'' subsets, which have roughly 20K images each. The COCO evaluation defines the object keypoint similarity (OKS) and uses the mean average precision (AP) over 10 OKS thresholds as main competition metric~\cite{COCOkeypoint}. The OKS plays the same role as the IoU in object detection. It is calculated from scale of the person and the distance between predicted points and GT points. Table~\ref{table:coco} shows results from top teams in the challenge.
It is noteworthy that our method has lower accuracy than the top-down methods on people of smaller scales ($\text{AP}^M$).
The reason is that our method has to deal with a much larger scale range spanned by all people in the image in one shot.
In contrast, top-down methods can rescale the patch of each detected area to a larger size and thus suffer less degradation at smaller scales.

\begin{table}[t]
\footnotesize
\begin{center}
\begin{tabular}{l |c| c c c c }
\hline
Team & \textbf{AP} & AP$^{50}$ &AP$^{75}$ &AP$^{M}$	&AP$^{L}$ \\
\hline
\multicolumn{6}{ c }{Test-challenge} \\
Ours &\textbf{60.5} &\textbf{83.4} &\textbf{66.4} &55.1 &\textbf{68.1} \\
G-RMI~\cite{papandreou2017towards} &59.8 &81.0 &65.1 &\textbf{56.7} &66.7 \\
DL-61 &53.3 &75.1 &48.5 &55.5 &54.8 \\
R4D &49.7 &74.3 &54.5 &45.6 &55.6 \\
\hline
\multicolumn{6}{ c }{Test-dev} \\
Ours &\textbf{61.8} &\textbf{84.9} &\textbf{67.5} &57.1 &\textbf{68.2} \\
G-RMI~\cite{papandreou2017towards} &60.5 &82.2 &66.2 &\textbf{57.6} &66.6 \\
DL-61 &54.4 &75.3 &50.9 &58.3 &54.3 \\
R4D &51.4 &75.0 &55.9 &47.4 &56.7  \\
\hline
\end{tabular}
\end{center}
\vspace{-5pt}
\caption{Results on the COCO 2016 keypoint challenge. Top: results on test-challenge. Bottom: results on test-dev (top methods only). AP$^{50}$ is for OKS $=0.5$, AP$^{L}$ is for large scale persons.}
\label{table:coco}
\vspace{-10pt}
\end{table}

\vspace{-5pt}
\begin{table}[h]
\footnotesize
\begin{center}
\begin{tabular}{l |c |c c c c }
\hline
Method & \textbf{AP} &	AP$^{50}$ &AP$^{75}$ &AP$^{M}$  &AP$^{L}$ \\
\hline
{\footnotesize GT Bbox + CPM~\cite{insafutdinov2016deepercut}} &62.7 &86.0 &69.3 &58.5 &70.6 \\ 
{\footnotesize SSD~\cite{liu2015ssd} + CPM~\cite{insafutdinov2016deepercut}} &52.7 &71.1 &57.2 &47.0 &64.2 \\
{\footnotesize Ours - 6 stages} &58.4 &81.5 &62.6 &54.4 &65.1\\
{\footnotesize \quad + CPM refinement} & 61.0 & 84.9& 67.5& 56.3& 69.3\\
\hline
\end{tabular}
\end{center}
\vspace{-5pt}
\caption{Self-comparison experiments on the COCO validation set.}
\vspace{-10pt}
\label{table:coco_val}
\end{table} 
 
\begin{figure*}[h]
\centering
  \includegraphics[width=1\linewidth]{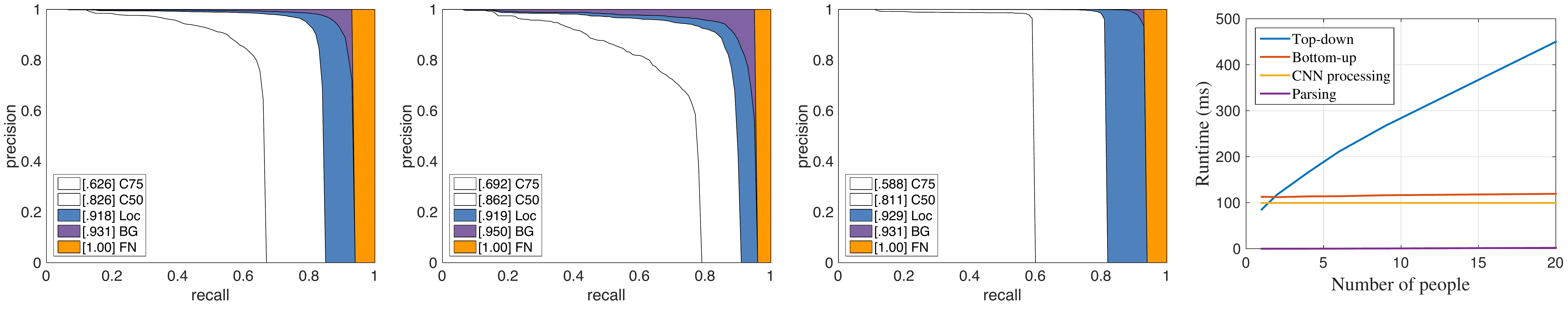}\\
  \vspace{-5pt}
  \hspace{10pt}{\footnotesize (a) PR curve - all people}\hspace{30pt} {\footnotesize (b) PR curve - large scale person} \hspace{15pt} {\footnotesize (c) PR curve - medium scale person} \hspace{30pt} {\footnotesize (d) Runtime analysis}\\
   \vspace{2pt}
  \caption{AP performance on COCO validation set in (a), (b), and (c) for Section~\ref{sec:coco}, and runtime analysis in (d) for Section~\ref{sec:runtime}.}
\label{fig:coco_ap}
 \end{figure*}

In Table~\ref{table:coco_val}, we report self-comparisons on a subset of the COCO validation set, i.e., 1160 images that are randomly selected. If we use the GT bounding box and a single person CPM~\cite{wei2016convolutional}, we can achieve a upper-bound for the top-down approach using CPM, which is $62.7\%$ AP. If we use the state-of-the-art object detector, Single Shot MultiBox Detector (SSD)\cite{liu2015ssd}, the performance drops $10\%$. 
%
This comparison indicates the performance of top-down approaches rely heavily on the person detector. In contrast, our bottom-up method achieves $58.4\%$ AP. If we refine the results of our method by applying a single person CPM on each rescaled region of the estimated persons parsed by our method, we gain an $2.6\%$ overall AP increase.
Note that we only update estimations on predictions that both methods agree well enough, resulting in improved precision and recall.
We expect a larger scale search can further improve the performance of our bottom-up method. Fig.~\ref{fig:coco_ap} shows a breakdown of errors of our method on the COCO validation set. 
Most of the false positives come from imprecise localization, other than 
background confusion. This indicates there is more improvement space in capturing spatial dependencies than in recognizing body parts appearances.

\begin{figure*}[h!]
\begin{center}
\includegraphics[width=1\linewidth]{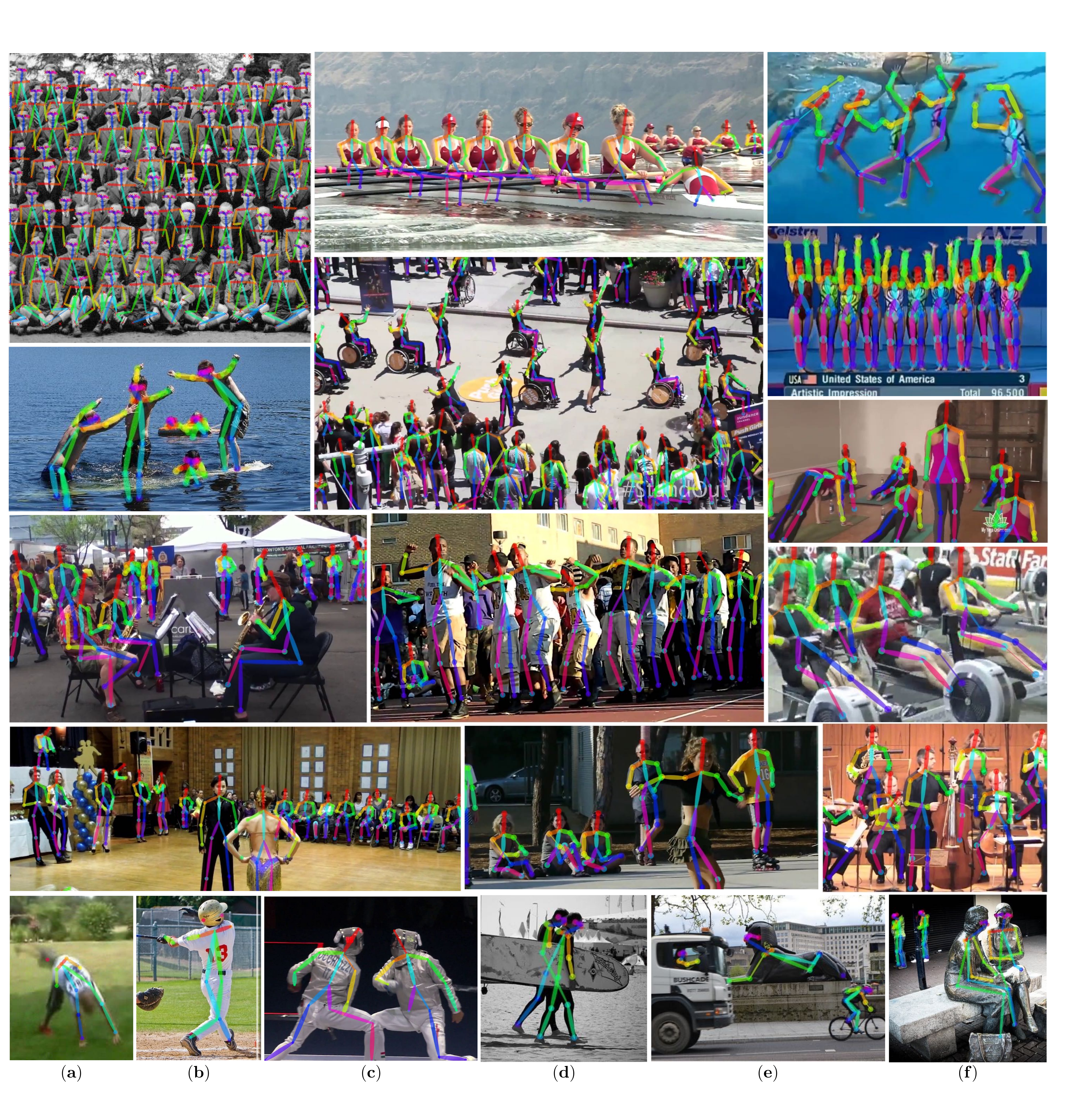}
\end{center}
\vspace{-10pt}
   \caption{Common failure cases: (a) rare pose or appearance, (b) missing or false parts detection, (c) overlapping parts, i.e., part detections shared by two persons, (d) wrong connection associating parts from two persons, (e-f): false positives on statues or animals.}
\vspace{-10pt}
\label{fig:failure}
\end{figure*}

\begin{figure*}[t]
\centering
\includegraphics[width=0.985\linewidth]{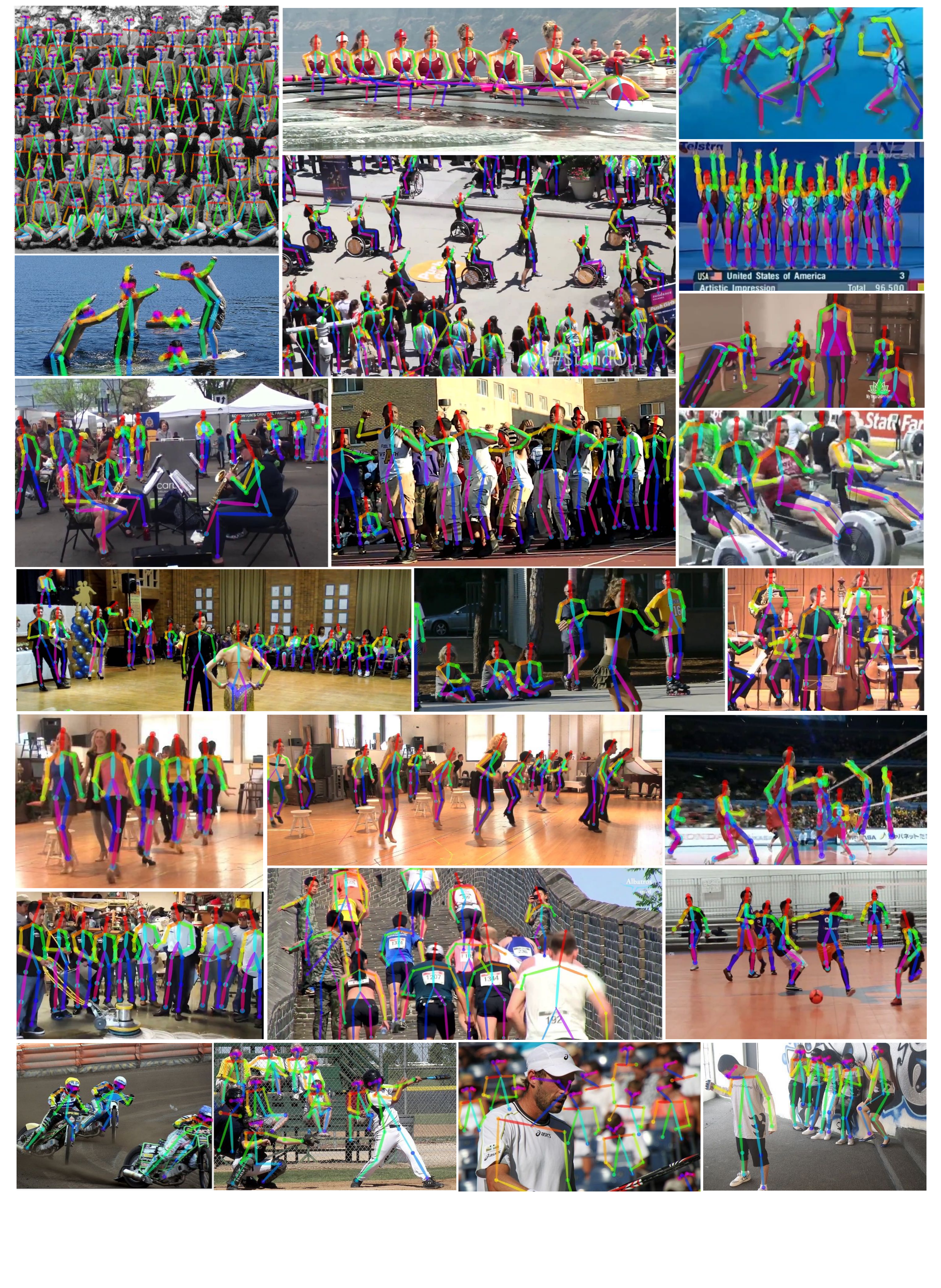}\\
\caption{Results containing viewpoint and appearance variation, occlusion, crowding, contact, and other common imaging artifacts. }
\vspace{-10pt}
\label{fig:qua}
\end{figure*}

\subsection{Runtime Analysis}\label{sec:runtime}

To analyze the runtime performance of our method, we collect videos with a varying number of people. The original frame size is $1080\times1920$, which we resize to $368\times654$ during testing to fit in GPU memory. The runtime analysis is performed on a laptop with one NVIDIA GeForce GTX-1080 GPU. 
In Fig.~\ref{fig:coco_ap}d, we use person detection and single-person CPM as a top-down comparison, where the runtime is roughly proportional to the number of people in the image. In contrast, the runtime of our bottom-up approach increases relatively slowly with the increasing number of people. The runtime consists of two major parts: (1) CNN processing time whose runtime complexity is $O(1)$, constant with varying number of people; (2) Multi-person parsing time whose runtime complexity is $O(n^2)$, where $n$ represents the number of people. However, the parsing time does not significantly influence the overall runtime because it is two orders of magnitude less than the CNN processing time, e.g., for $9$ people, the parsing takes $0.58$ ms while CNN takes $99.6$ ms. Our method has achieved the speed of $8.8$ fps for a video with $19$ people.

\section{Discussion}
Moments of social significance, more than anything else, compel people to produce photographs and videos. Our photo collections tend to capture moments of personal significance: birthdays, weddings, vacations, pilgrimages, sports events, graduations, family portraits, and so on. To enable machines to interpret the significance of such photographs, they need to have an understanding of people in images. Machines, endowed with such perception in realtime, would be able to react to and even participate in the individual and social behavior of people. 

In this paper, we consider a critical component of such perception: realtime algorithms to detect the 2D pose of multiple people in images. We present an explicit nonparametric representation of the keypoints association that encodes both position and orientation of human limbs. Second, we design an architecture for jointly learning parts detection and parts association. Third, we demonstrate that a greedy parsing algorithm is sufficient to produce high-quality parses of body poses, that maintains efficiency even as the number of people in the image increases. We show representative failure cases in Fig.~\ref{fig:failure}. We have publicly released our \href{https://github.com/ZheC/Realtime_Multi-Person_Pose_Estimation}{code} (including the trained models) to ensure full reproducibility and to encourage future research in the area. 

\section*{Acknowledgements}
We acknowledge the effort from the authors of the MPII and COCO human pose datasets. These datasets make 2D human pose estimation in the wild possible. This research was supported in part by ONR Grants N00014-15-1-2358 and N00014-14-1-0595.

{\small
\bibliographystyle{ieee}
\bibliography{egbib}
}

\end{document}